%% file: main.tex
\documentclass[lettersize,journal]{IEEEtran}
\usepackage{amsmath,amsfonts}
\usepackage{algorithmic}
\usepackage{algorithm}
\usepackage{array}
\usepackage[caption=false,font=normalsize,labelfont=sf,textfont=sf]{subfig}
\usepackage{textcomp}
\usepackage{stfloats}
\usepackage{url}
\usepackage{verbatim}
\usepackage{graphicx}
\usepackage{cite}
\usepackage{xcolor}
\usepackage{multirow}
\begin{document}

\title{A Hardware-oriented Approach for Efficient Bayesian Inference Computation and Deployment}



\author{
Nikola~Pižurica\textsuperscript{2,4},
Matteo~Risso\textsuperscript{3},
Nikola~Milović\textsuperscript{4},
Alessio~Burrello\textsuperscript{3},
Igor~Jovančević\textsuperscript{2,4},
Conor~Heins\textsuperscript{5},
and~Miguel~de~Prado\textsuperscript{1}%
\thanks{\textsuperscript{1}PRAESC, Biel, Switzerland}%
\thanks{\textsuperscript{2}Computer Science Center,
University of Montenegro}%
\thanks{\textsuperscript{3}Department of Control and Computer Engineering,
Politecnico di Torino}%
\thanks{\textsuperscript{4}Fain Tech, Podgorica, Montenegro}%
\thanks{\textsuperscript{5}Independent Researcher}%
\thanks{Corresponding author: Nikola Pižurica
(e-mail: nikola.p@ucg.ac.me).}

\thanks{This work has been submitted to the IEEE for possible publication. Copyright may be transferred without notice, after which this version may no longer be accessible.}

}

\markboth{Journal of \LaTeX\ Class Files,~Vol.~14, No.~8, August~2021}%
{Shell \MakeLowercase{\textit{et al.}}: A Sample Article Using IEEEtran.cls for IEEE Journals}


\maketitle

\begin{abstract}
Bayesian inference provides a principled foundation for reasoning under uncertainty, but its computational cost hinders deployment on resource-constrained edge devices. In this paper, we present a hardware-oriented methodology for accelerating discrete Bayesian inference on commercial off-the-shelf embedded GPUs. We identify that the latency of a broad class of variational message-passing algorithms is dominated by tensor contractions. Our approach restructures the memory layout of these operations using two complementary merging strategies that produce compact, regularly-shaped primitives better suited for efficient GPU execution. We then introduce optional sparse array representations and a tensor-clustering scheme to reduce the memory footprint. We instantiate the methodology and produce optimized variants of three message-passing algorithms for Hidden Markov Models (HMMs), namely variational filtering, variational message passing, and marginal message passing. Furthermore, we complement this with a machine-learning-based autotuner that automatically selects the best-performing algorithmic variant for a given generative model specification. Benchmarked on an NVIDIA Jetson Orin AGX across $770$ randomly sampled realistic Partially Observable Markov Decision Process (POMDP) configurations, our implementations achieve speedups of up to 5x, with typical gains of 2--2.5x, while producing numerically identical outputs to the baseline implementations.

\end{abstract}


\section{Introduction}

Uncertainty is an inherent feature of the real world. All living beings continuously face incomplete and noisy sensory information about their environment, yet they must make timely decisions to survive. Thus, the ability to reason under uncertainty is a core requirement for adaptive behavior. A substantial body of work in theoretical neuroscience suggests that the brain itself performs probabilistic inference over the latent causes of sensory inputs~\cite{friston2010FEP}. Artificial intelligent systems deployed in the real world face fundamentally analogous challenges, operating under partial observability, sensor noise, and environmental ambiguity, where incorrect estimates can lead to costly failures. Bayesian inference provides a principled mathematical foundation for uncertainty-aware reasoning, enabling agents to maintain and update their probabilistic beliefs over latent variables as new evidence arrives. Various Bayesian approaches have found practical applications across many domains, including robotics~\cite{robotics1, pezzato2025mobile}, automotive applications~\cite{automotive1, automotive2}, recommendation systems~\cite{stern2009matchbox, salakhutdinov2008bayesian}, healthcare~\cite{smith2021predictive, smith2022broken}, and the gaming industry~\cite{herbrich2006trueskill}.

Despite these advances, the computational cost of Bayesian inference remains a persistent challenge~\cite{angelino2016patterns}. Variational inference (VI) is a widely used instance of Bayesian inference that frames posterior approximation as an optimization problem, alleviating the original computational intractability. However, in practical VI implementations for probabilistic models, often represented as discrete structured factor graphs, belief update algorithms such as variational filtering, Variational Message Passing (VMP), and Marginal Message Passing (MMP)~\cite{parr2019neuronal}, rely on many computational operations expressible as generalized Einstein summations or tensor contractions~\cite{tensor_contractions} (e.g., marginalization steps, expected value calculations, and variable eliminations). The complexity of these operations scales polynomially with the number of random variables, making them increasingly expensive as models get deployed in richer, more realistic environments. This creates a trade-off between model capability and computational feasibility.

For many artificial intelligent agents, edge devices are the most natural, and often the only viable deployment setting. Mobile robots, autonomous vehicles, and embedded cyber-physical systems must perform inference locally, as cloud offloading is rarely an option due to latency constraints, unreliable connectivity, and data privacy concerns. Bayesian methods have shown considerable promise in these domains, from body perception and solving long-horizon tasks on robotic platforms~\cite{robotics1, pezzato2025mobile} to modeling uncertainty-aware driving behavior for automotive applications~\cite{automotive1, automotive2}. However, such demonstrations have largely been evaluated in simulations rather than real environments. The efficiency bottleneck becomes decisive in bridging this gap: running inference on embedded hardware imposes tight resource constraints, sharpening the accuracy–efficiency trade-off and demanding inference pipelines that are simultaneously precise and fast.

\looseness=-1
In this paper, we demonstrate how to leverage off-the-shelf embedded GPUs to accelerate and improve the efficiency of Bayesian inference algorithms. Our approach is based on memory layout transformations of the computational primitives that dominate inference runtime, specifically tensor contractions or Einstein summations, providing hardware-aware alternatives optimized for GPU execution. Our custom implementations improve the regularity of input shapes, which leads to better GPU mapping and faster execution. By focusing on generic computational primitives rather than the specifics of individual algorithms, we show that our methodology generalizes naturally, applying the same optimization strategy across multiple algorithms that involve tensor contractions, resulting in significant and consistent runtime reductions, all while remaining mathematically equivalent to the original implementations (i.e., in a lossless manner). This is showcased on three algorithms: variational Hidden Markov Model filtering (referred hereafter as the FPI algorithm due to its use of \textbf{f}ixed \textbf{p}oint \textbf{i}terations for optimization), as well as variational message passing (VMP) and marginal message passing (MMP).

The main contributions of this paper are as follows:

\begin{itemize}
    \item \textbf{A unified optimization methodology for Bayesian inference primitives that exploits GPU parallelization.} We propose a systematic approach for accelerating tensor contraction operations that arise in discrete Bayesian inference. This methodology is algorithm-agnostic and applicable to a wide class of inference procedures.

    \item \textbf{An optimized set of primitives that demonstrate algorithmic acceleration for general Bayesian inference, e.g., FPI, VMP, and MMP.} We deliver GPU-accelerated variants that achieve speedups of up to 5x over the baseline implementations, with typical gains in the range of 2--2.5x. Moreover, our implementations do not rely on additional approximations and therefore produce exactly the same outputs as the baselines.

    \item \textbf{ML-based, hardware-aware autotuner for configuration selection.} We introduce an autotuning module that uses a trained predictor to automatically select the fastest inference configuration for a given generative model, eliminating the need for exhaustive benchmarking during deployment. Such benchmark processes take around 4 minutes on average, and 20--25 minutes for large generative models, whereas the autotuner delivers highly accurate predictions in a fraction of a second.

    \item \textbf{Open-source release.} Our code will be made publicly available. Link omitted for the double-blind review.
\end{itemize}

\section{Background and Related Work}

\subsection{Bayesian and variational inference}
 
Bayesian inference provides the formal substrate for reasoning under uncertainty, but exact posteriors are typically intractable in models of practical interest. Two broad families of approximate inference have emerged in response. Variational inference (VI) reformulates posterior estimation as the optimization of a surrogate distribution (the variational distribution) by minimizing a bound on the marginal likelihood, which when minimized also brings the surrogate distribution as close as possible to the true posterior \cite{jordan1999introduction, blei2017variational}, while Markov chain Monte Carlo (MCMC) and related sampling methods characterize the posterior through draws obtained from carefully constructed stochastic processes. Within VI, message-passing variants such as variational message passing~\cite{winn2005variational}, expectation propagation~\cite{exp_prop}, and belief propagation~\cite{bel_prop} on factor graphs underlie many practical implementations of Bayesian inference in graphical models, and form the algorithmic core of the message-passing routines that we accelerate in this work~\cite{dacosta2020synthesis}. 
 
Variational and message-passing methods have a long history of deployment in the real world. Bayesian skill rating in commercial multiplayer gaming~\cite{herbrich2006trueskill}, large-scale Bayesian matrix factorization for movie recommendation on Xbox~\cite{stern2009matchbox} and Netflix~\cite{salakhutdinov2008bayesian}, and Bayesian collaborative filtering at scale~\cite{liang2018variational} have all relied on variational message passing or amortized variational inference to scale Bayesian computations to industrial workloads. Additionally, Kalman filters and their nonlinear extensions remain standard components of aerospace, navigation, and sensor-fusion stacks~\cite{linderman2025dynamax, gtsam}.

A particularly promising recent line of research is active inference~\cite{friston2016activelearning, ActInfBook}, which casts both perception and action as the minimization of a single variational free-energy objective over a generative model of the environment. It has been carried from theory into deployed systems across several domains. In computational psychiatry, it serves as a normative model of symptoms in conditions such as depression, anxiety, and substance use disorders, to support diagnostics and treatment selection~\cite{smith2021predictive, smith2022broken}. In robotics, it underpins adaptive control of high-degree-of-freedom manipulators~\cite{robotics2}, body perception, self/other distinction, and reaching on humanoid platforms~\cite{robotics1, lanillos2021survey, oliver2021empirical}, as well as simultaneous mapping, localization, and exploration on mobile robots~\cite{catal2021navigation}. In the automotive industry, it has been used to model and evaluate driver--vehicle interactions~\cite{automotive1, automotive2}, suggesting a path towards ADAS-style decision support.

\vspace{-0.1cm}
\subsection{Computational bottlenecks and edge deployment}

Probabilistic programming frameworks such as Stan~\cite{carpenter2017stan}, Pyro~\cite{bingham2019pyro}, and NumPyro~\cite{phan2019numpyro} have made Bayesian algorithms broadly accessible. However, scaling up these methods to model increasingly complex scenarios under strict latency budgets remains a bottleneck, particularly when inference must run alongside perception and control on edge devices.

\looseness=-1
Within this landscape, our work focuses specifically on the optimization under tight constraints of \emph{variational} inference over \emph{discrete} generative models, for two reasons. First, sampling-based methods such as MCMC and sequential Monte Carlo have long been regarded as computationally expensive and slow to converge. They characterize posteriors via long chains of stochastic draws whose convergence is data-dependent and difficult to bound a priori, making them poorly suited for deployment on edge devices with limited computational resources. Message-passing algorithms, by contrast, cast inference as deterministic optimization over a bounded number of iterations, yielding a predictable per-step runtime that can exploit GPU parallelism. Second, we focus on discrete generative models due to the prevalence of discrete probabilistic agents in real-world edge applications and the growing popularity of software frameworks tailored for categorical state spaces \cite{heins2022pymdp, zhou2024pgmax, aifjl, spm12_software}. Thus, targeting such cases addresses a concrete and widely shared deployment challenge in the Bayesian inference community. Furthermore, discrete generative models admit exact closed-form variational updates expressed as tensor contractions over factorized parameter arrays, whose structural regularity can be exploited by SIMD/SIMT hardware architectures~\cite{peng2024evaluating} present in GPUs, TPUs, and NPUs.
 
Industrial Bayesian inference applications typically require high-dimensional generative models, batched inference over multiple models and observations, and meeting strict latency constraints, all of which expose the cost of repeated tensor contractions and message updates. Thanks to its Python frontend, \texttt{pymdp}~\cite{heins2022pymdp} has democratized access to discrete active inference while also enabling GPU execution via the JAX backend. RxInfer.jl~\cite{bagaev2023rxinfer} delivers reactive message passing on factor graphs in Julia, PGMax~\cite{zhou2024pgmax} implements loopy belief propagation for discrete graphical models on top of JAX, and Dynamax~\cite{linderman2025dynamax} provides JAX-accelerated inference for state-space models. Collectively, these efforts confirm that message-passing inference benefits substantially from modern array-programming backends. However, their performance on any given device remains tied to the low-level code emitted by the underlying compiler, and hardware-aware optimization on embedded platforms has received comparatively little attention. A small but growing literature targets specialized hardware for probabilistic inference, including FPGA accelerators for Bayesian Confidence Propagation Networks and spiking-Bayesian inference~\cite{ravichandran2024bcpnnfpga}. Our work complements this line by focusing on commodity-embedded GPUs (NVIDIA Jetson family) and the message-passing primitives shared by many inference algorithms, demonstrating that significant speedups are achievable without departing from the standard Python/JAX deployment path.

\section{Methodology}
\label{sec:methodology}

\subsection{Main computational primitives in Bayesian inference}

Across a wide class of VI algorithms, the dominant computational workload arises from repeatedly evaluating expectations of log-factors under partially factorized distributions. In discrete-state-space models, these expectations reduce to tensor contraction operations between factor potentials and marginal beliefs. We therefore identify tensor contraction as the fundamental computational primitive underlying Bayesian inference in these settings, and focus our optimization efforts at this level.

More concretely, consider a variational free energy objective over hidden states $\mathbf{s} = \{s^1, s^2, \dots, s^F\}$ under a factorized posterior approximation $q(\mathbf{s}) = \prod_{f=1}^F q(s^f)$. Coordinate-wise optimization of this objective yields update equations of the generic form \cite{winn2005variational, blei2017variational}
\[
\log q(s^f) \propto \mathbb{E}_{q(\mathbf{s}^{\setminus f})}\left[\log p(\mathbf{o}, \mathbf{s})\right],
\]
where $\mathbf{s}^{\setminus f}$ denotes all hidden state variables except $s^f$. This means that updating any one posterior factor requires computing an expected log-energy with respect to the remaining marginals. When the generative model factorizes into observation and transition terms, these updates decompose naturally into likelihood, prior, and temporal message components. In temporal models such as HMMs, analogous quantities also appear in the forward and backward messages used to propagate information across time. In discrete settings, these expected log-energy terms can be written as contractions between high-dimensional parameter tensors and lower-dimensional posterior marginals. Although the exact quantities may arise in different inference algorithms and under different message scheduling schemes, they all share the same underlying computational form: repeated Einstein summations over heterogeneous tensor shapes.

This finding motivates the central perspective of our work. We treat various Bayesian algorithms as distinct inference procedures built from a common set of message-passing primitives. Instead of optimizing each algorithm in isolation, our methodology targets the low-level contraction operators that instantiate the messages and algorithms, aiming to improve inference runtime in a lossless manner (i.e., preserving the  numerical equivalence with respect to the original inference updates).

\vspace{-0.1cm}
\subsection{Terminology and notation}


Although our methodology applies to many Bayesian inference settings, we primarily demonstrate it for factorized observations, hidden states, and controls. Such HMMs are often used as generative models in Partially-Observable Markov Decision Process (POMDP) setups~\cite{kaelbling1998pomdp}. Letting $\mathbf{o}_{1:T} = (\mathbf{o}_1, \dots, \mathbf{o}_T)$ and $\mathbf{s}_{1:T} = (\mathbf{s}_1, \dots, \mathbf{s}_T)$ denote sequences of observations and hidden states over $T$ time steps, and $\mathbf{u}_{1:T-1} = (\mathbf{u}_1, \dots, \mathbf{u}_{T-1})$ a sequence of controls, the joint distribution factorizes as $p(\mathbf{o}_{1:T}, \mathbf{s}_{1:T} \mid \mathbf{u}_{1:T-1};\, \mathbf{A}, \mathbf{B})
\;=\;
p(\mathbf{s}_1)\,
\prod_{t=2}^{T} p(\mathbf{s}_t \mid \mathbf{s}_{t-1}, \mathbf{u}_{t-1};\, \mathbf{B})\,
\prod_{t=1}^{T} p(\mathbf{o}_t \mid \mathbf{s}_t;\, \mathbf{A})$, where the emission model $p(\mathbf{o}_t \mid \mathbf{s}_t; \mathbf{A})$ and the transition model $p(\mathbf{s}_t \mid \mathbf{s}_{t-1}, \mathbf{u}_{t-1}; \mathbf{B})$ are parameterized by the tensor families $\mathbf{A} = \{A_1, \dots, A_M\}$ and $\mathbf{B} = \{B_1, \dots, B_F\}$, respectively. Observations, hidden states, and controls are all assumed to admit a factorization into multiple discrete variables: at each time step $t$, $\mathbf{o}_t = \{o_t^1, o_t^2, \dots, o_t^M\}$ comprises $M$ observation modalities, $\mathbf{s}_t = \{s_t^1, s_t^2, \dots, s_t^F\}$ comprises $F$ hidden state factors, and $\mathbf{u}_t = \{u_t^1, u_t^2, \dots, u_t^F\}$ contains one control variable for each hidden state factor.

Different inference algorithms operate on different temporal windows of this model. In algorithms that consider the current time step in isolation, only $\mathbf{o}_t$ and $\mathbf{s}_t$ at a single moment $t$ are relevant. In settings that require longer temporal context, inference is performed over a window comprising the current time step $t$ together with the preceding $h$ time steps, denoted $\mathbf{o}_{t-h:t} = \{o_{t-h:t}^1, \dots, o_{t-h:t}^M\}$ and $\mathbf{s}_{t-h:t} = \{s_{t-h:t}^1, \dots, s_{t-h:t}^F\}$. The parameter $h$ is from now on referred to as the \textit{inference horizon}.

Each observation modality $o^i$ takes on one of $m_i$ values, also referred to as levels in the literature, and is represented as a categorical probability vector of length $m_i$. Similarly, each hidden state factor $s^j$ takes on one of $n_j$ levels and is represented by a vector of length $n_j$. When using sequences of observations and hidden states (e.g., in the MMP and VMP algorithms), the per-time-step probability vectors form matrices of shapes $(h+1) \times m_i$ and $(h+1) \times n_j$, respectively. When working with batches of differently parameterized generative models of batch size $b$, these become 3D arrays of shapes $b \times (h+1) \times m_i$ and $b \times (h+1) \times n_j$.

The A-arrays introduced above admit a more detailed structure of the emission model. Each array $A_i$ represents the conditional probability distribution $P(o^i|s^{i_1}, s^{i_2}, ..., s^{i_{k_i}})$, and the observation modality $o^i$ is said to depend only on some $k_i$ hidden state factors $s^{i_1}, s^{i_2}, ..., s^{i_{k_i}}$, not necessarily all possible ones. In the most general case, the shape of array $A_i$ would be $b \times m_i \times n_{i_1} \times n_{i_2} \times ... \times n_{i_{k_i}}$. The lists of hidden state factors that each observation modality depends on are often referred to as A-dependencies. Our methodology for optimizing state inference efficiency can handle completely arbitrary A-dependencies.

In a similar manner to A-arrays, each $B_j$ corresponds to a conditional probability distribution $P(s_{t+1}^j|s_t^{j_1}, s_t^{j_2}, ..., s_t^{j_{k_j}}, u_t^j)$. The hidden state factor $s^j$ depends on some $k_j$ hidden state factors $s^{j_1}, s^{j_2}, ..., s^{j_{k_j}}$ (which are usually specified as a list of B-dependencies), as well as some control state $u^j$. Contrary to A-dependencies, the B-dependency lists are often trivial, i.e., having each hidden state factor at time step $t+1$ depend only on its own value at time step $t$: $B_j = P(s_{t+1}^j|s_t^j, u_t^j)$. The proposed optimization methodology currently supports only trivial B-dependencies, but this can cover a significant portion of practical use cases. Nonetheless, we plan to extend our approach to arbitrary B-dependencies in our future work.

\subsection{Optimizing tensor contractions}
\label{sec:optimizing_contractions}

\begin{figure}[htbp]
\centering
\includegraphics[width=3.15in]{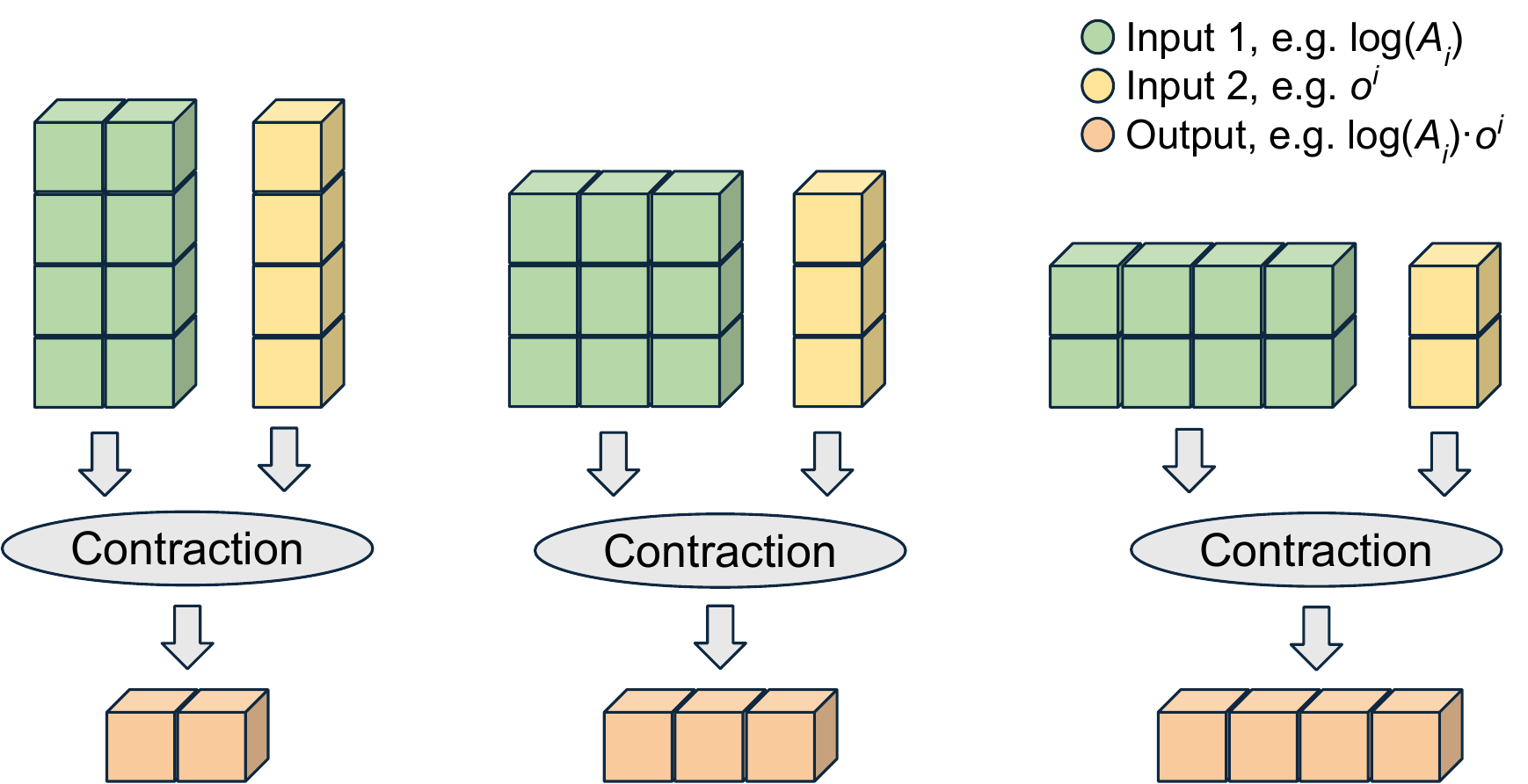}
\caption{A visualization of contraction operations over different inputs. Only 1D and 2D arrays are shown for simplicity, but the concept generalizes easily to higher-dimensional inputs as well.}
\label{fig:original_computation}
\end{figure}

In various Bayesian inference applications, one contraction operation often needs to be performed for a collection of input arrays, as illustrated in Fig.~\ref{fig:original_computation}. This is usually implemented by looping the same computational primitive over different inputs. A typical example, used as a running illustration throughout this paper, is the log-likelihood computation common to all state inference algorithms considered here: the logarithm of each A-array $A_i$ (green) is multiplied by the corresponding observation vector $o^i$ (yellow) along the observation axis, which is then summed out (orange). Since this generalizes matrix-vector multiplication, we denote it as $\log(A_i) \cdot o^i$.

\vspace{-0.05cm}

Performing contractions in this looped form can be inefficient. The ranks and shapes of input arrays can vary significantly across the collection, which generally leads to poor GPU mapping due to the following reasons.

\begin{itemize}
    \item Looping numerical operations over separate arrays causes a kernel launch overhead that can grow significantly large, especially when aggregated over many iterations.
    \item Feeding the GPU one pair of input arrays at a time results in suboptimal utilization of available memory bandwidth, leading algorithms to operate in memory-bound rather than compute-bound conditions.
    \item Variations in input shapes often lead to underutilization of parallelism (e.g., it is impossible to simply stack the inputs and perform operations in vectorized form). Furthermore, the fusion of low-level operations is hindered.
\end{itemize}

To address these difficulties, we propose two strategies for padding and merging the inputs prior to contraction. The goal is to replace a loop of heterogeneous, small operations with a single, compact operation that maps cleanly to GPU hardware. We outline both strategies conceptually below, and the precise construction of the merged arrays is given in the Appendix. It is important to emphasize that padding and merging are performed only once during initialization and, as such, do not incur computational overhead at inference time.

\subsubsection{Axis-aligned padding and merging} 
\begin{figure*}[htbp]
\centering
\subfloat[]{\includegraphics[width=2.25in]{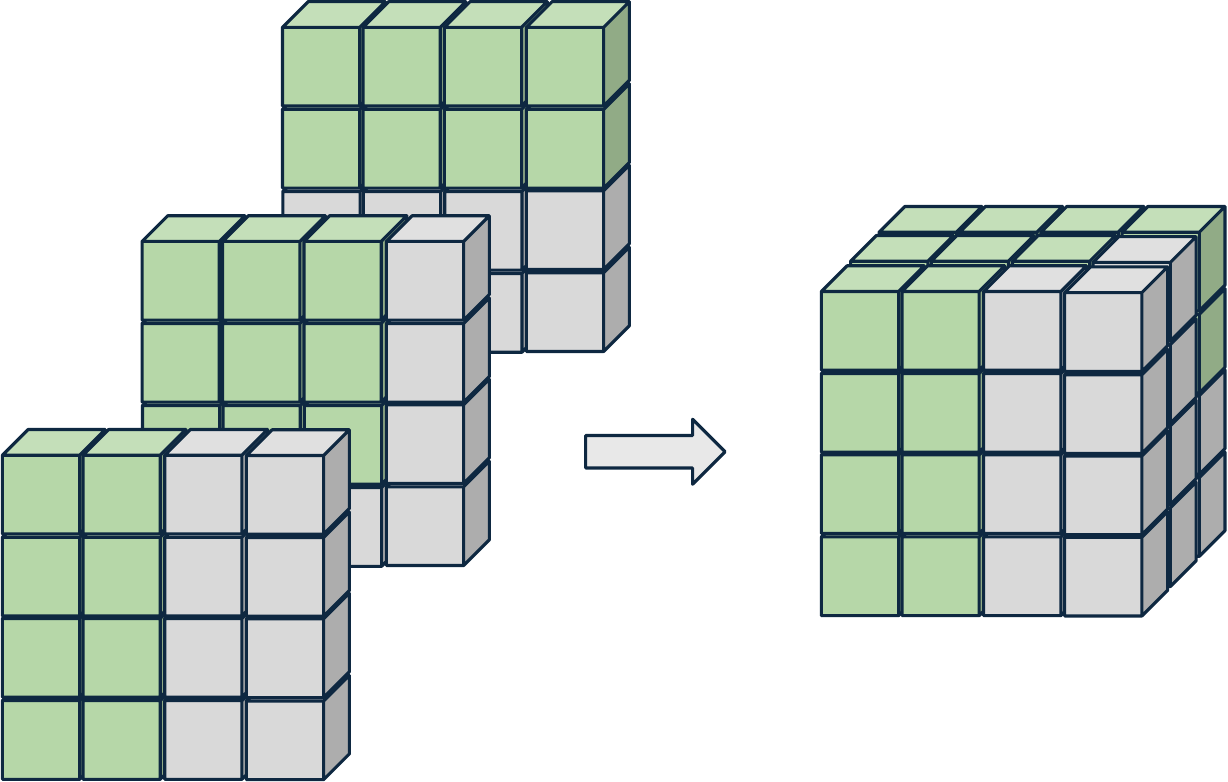}%
\label{fig:padding_and_merging}}
\hfil
\subfloat[]{\includegraphics[width=1.26in]{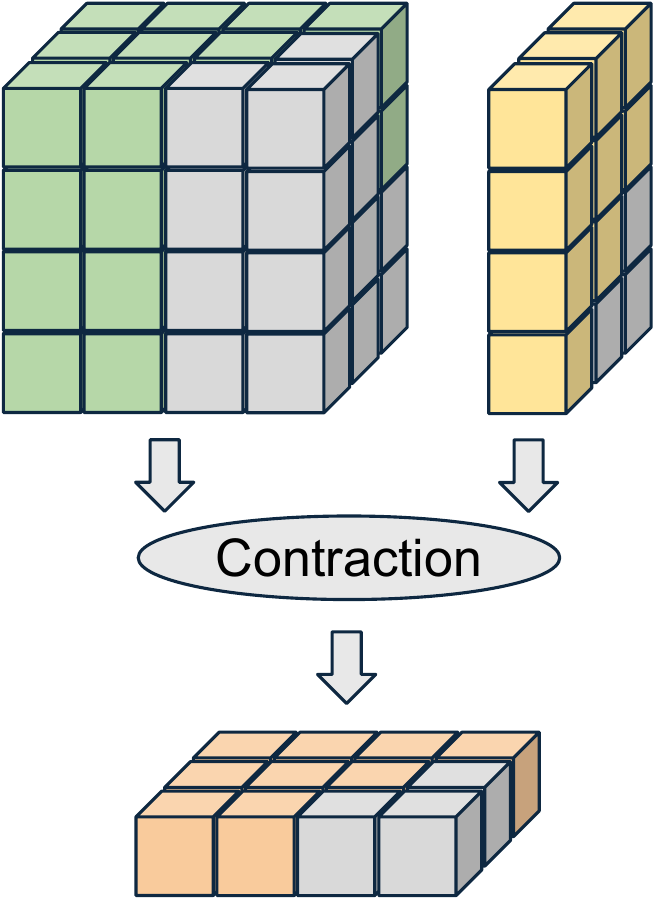}%
\label{fig:merged_computation}}
\hfil
\subfloat[]{\includegraphics[width=2.07in]{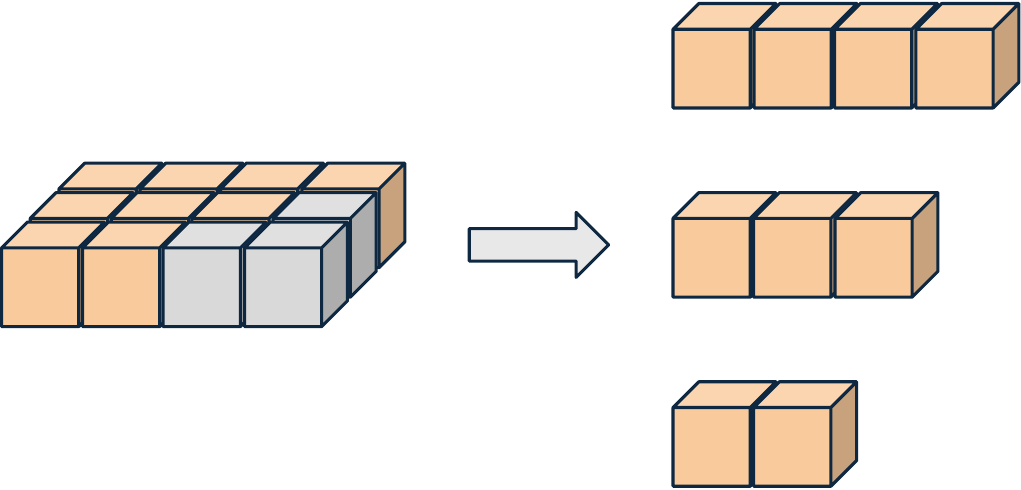}%
\label{fig:deconstruction}}
\caption{(a) A visualization of the axis-aligned padding and merging strategy. Gray elements represent zero values that are added for padding. (b) Performing a tensor contraction in axis-aligned merged form. (c) Deconstructing the output of a merged tensor contraction into 
individual outputs.}
\label{fig_sim}
\end{figure*}

The first approach is to zero-pad each A-array up to a common shape and concatenate the results along the batch axis into a single tensor $\mathbf{\bar{A}}$ (Fig.~\ref{fig:padding_and_merging}). Analogously, the observation vectors $o^i$ are also merged into a single tensor $\mathbf{\bar{o}}$, which is then broadcast to match the rank of $\mathbf{\bar{A}}$. The tensor contraction can then be performed in merged form as a single broadcasting multiplication between $\log(\mathbf{\bar{A}})$ and $\mathbf{\bar{o}}$, followed by a summation over the observation axis. This is illustrated in Fig.~\ref{fig:merged_computation}. The merged output is a zero-padded concatenation of what the looped version would produce, and can be split back into individual outputs when needed (Fig.~\ref{fig:deconstruction}). Despite its simplicity, this strategy can lead to substantial speedups. Its main drawback is a tendency to an explosion in memory requirements when the input arrays differ greatly in rank or shape, which we address in Section~\ref{sec:memory}.

\subsubsection{Block-diagonal merging} 
\begin{figure}[htbp]
\centering
\includegraphics[width=2in]{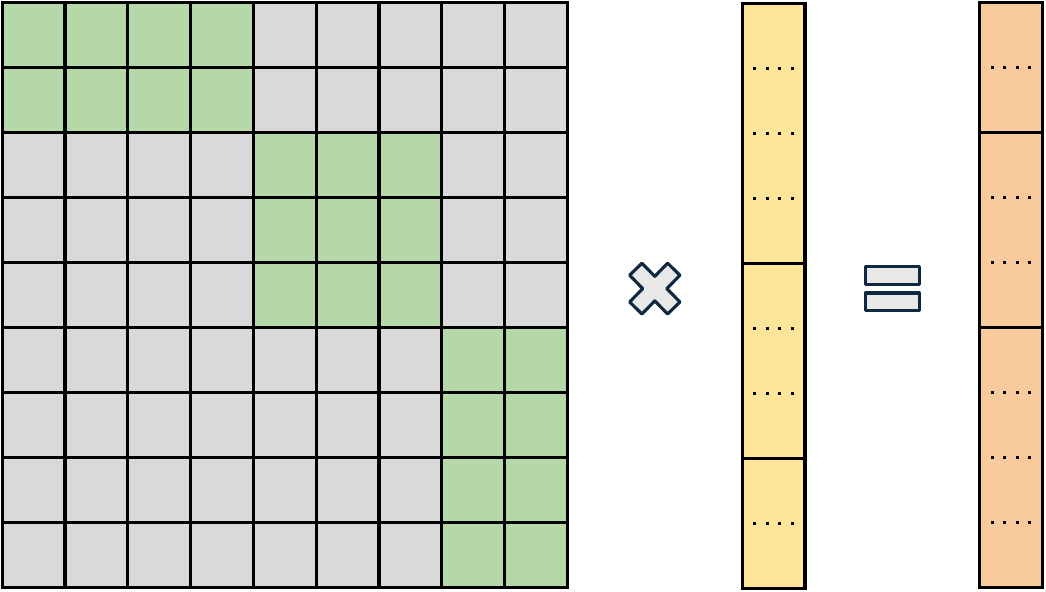}
\caption{Merging tensor contractions into one standard matrix-vector multiplication. High-dimensional inputs (e.g., A-arrays) are arranged into a block-diagonal matrix. \vspace{-0.1cm}}
\label{fig:block_computation}
\end{figure}

The second strategy is tailored to settings where one of the two operands is one-dimensional, and the other has an arbitrary rank (a pattern that naturally fits the log-likelihood computation). The high-rank inputs are reshaped so that all of their lagging dimensions are flattened, and the resulting matrices are arranged as blocks of a single block-diagonal matrix $\mathbf{A_{block}}$. The one-dimensional inputs are concatenated into a long vector $\mathbf{o_{concat}}$. The entire loop of contractions then reduces to a standard matrix-vector multiplication $\log(\mathbf{A_{block}}) \cdot \mathbf{o_{concat}}$, the output of which can again be split and reshaped into the individual per-input results. The idea is illustrated in Fig.~\ref{fig:block_computation}.

Both strategies trade a small amount of redundant computation (on 
padded or zero-valued entries) for substantially better hardware 
utilization.

\vspace{-0.2cm}
\subsection{Reducing the memory requirements of merged contractions}
\label{sec:memory}

A drawback of the previously described methods is that the sizes of merged arrays grow rapidly as the number of input arrays increases. This can lead to unfeasible memory requirements, which need to be accounted for. Moreover, the overhead of loading extremely large arrays and performing excessive computation on padded zeros can neutralize the benefits of improved GPU utilization.

The first approach we employ in this study leverages efficient memory representation of sparse arrays and basic numerical operations on them. The reasoning behind this is simple - the padding and merging strategies produce large arrays that are sometimes mostly filled with artificial zeros. We utilize the batched coordinate (BCOO) sparse representation \cite{sparse_formats} of JAX arrays in our experiments.

In addition to applying the existing JAX-BCOO functionalities, we propose a method of our own to alleviate memory-related challenges. Instead of merging all tensor contractions into a single contraction, it is possible to divide them into groups (clusters) and perform per-cluster merging. If clusters are chosen advantageously, for example, by grouping similarly shaped arrays, we can drastically reduce the number of artificial zeros used for padding while still retaining the hardware-related benefits of merged computations. Since clustering precedes array padding and merging, it is also a one-of procedure that does not incur computational overhead at inference time. As before, we explain this concept on the log-likelihood example, using the already established notation.

For the axis-aligned merging strategy, we propose a rank-based clustering of input arrays. Suppose that the lengths of A-dependencies can vary from $1$ to some upper limit $k$. As a consequence, the ranks of arrays $\{A_1, A_2, ..., A_M\}$ would vary from $3$ to $k+2$ in the most general batched form. All A-arrays sharing the same rank $r$ form one cluster, which we will denote as $\{A_1^r, A_2^r, ..., A_{M_r}^r\}$. These arrays are merged into $\mathbf{\bar{A}_r}$ using the previously explained methodology. We follow the same steps, only limited to the set $\{A_1^r, A_2^r, ..., A_{M_r}^r\}$, instead of taking all A-arrays $\{A_1, A_2, ..., A_M\}$ into consideration. Repeating this across all ranks yields a set of merged computations that balance two extremes: excessively looping over individual input arrays, or merging everything into a single excessively large array.

The block-diagonal merging strategy requires a different clustering approach, as the rank-based method would not be meaningful given the flattening of lagging dimensions. We propose an approach that is reminiscent of agglomerative clustering. Initially, each A-array is treated as a trivial block-diagonal matrix (with no artificial zero elements), i.e. $\mathbf{A_{block}^i} = A_i$. Then, an iterative procedure is performed, in which we repeatedly select two block-diagonal matrices $(\mathbf{A_{block}^i}, \mathbf{A_{block}^j})$ whose merging will increase the number of artificial zeros by the least amount. Supposing that dimensions of the two selected matrices are $r_i \times c_i$ and $r_j \times c_j$, it is easy to express the increase in zeros used for padding algebraically: $\Delta_z = (r_i + r_j)(c_i + c_j) - (r_i c_i + r_j c_j) = r_i c_j + r_j c_i$. Upon determining the minimal $\Delta_z$, the corresponding pair of matrices is merged into a new block-diagonal matrix. By this construction, each new iteration leads to a larger increase in the total number of artificial zeros, and, if plotted, these numbers form a hyperbolic curve. We empirically found that the set of block-diagonal matrices corresponding to the elbow of the curve generally yields the best results (faster inference due to higher unification of tensor contractions, while keeping the number of padded zeros reasonably low).

\vspace{-0.2cm}
\section{Implementation}
\label{sec:variants}

From a message-passing viewpoint, discrete state inference is generally assembled from likelihood messages (arising from the observation model, i.e., from A-arrays) and transition messages (arising from the state dynamics model, i.e., from B-arrays). Computing likelihood messages involves two contraction-heavy steps: the log-likelihood computation -- used as a running example throughout this paper -- and a marginalization step that evaluates likelihood expectations under the current posterior approximations and aggregates them through the Markov blanket principle~\cite{pearl1988probabilistic}. Since our methodology supports arbitrary A-dependencies, this process generally involves operations with high-rank arrays. Conversely, transition messages are computed in a simpler manner --- they mainly involve contractions between low-rank B-arrays (due to trivial B-dependencies) and current posterior approximations. 

Due to this essential difference in complexity, our methodology handles the two types of messages differently. We limit the computation of transition messages to the plain axis-aligned merged form (admissible due to the low-rank nature of B-arrays and the lack of memory-related concerns). On the other side, the likelihood messages offer several optimization approaches depending on how they are computed, yielding the following families of optimized implementations.

\emph{1) Hybrid} methods perform only the log-likelihood computation in merged form, immediately deconstruct the result into per-modality outputs, and delegate the marginalization step to the original, looped implementation. The name reflects the combination of our merging strategies with the original logic.

\emph{2) End-to-end} methods keep the merged representation throughout the entire computation of likelihood messages, including marginalization. This requires axis-aligned padding specifically, with one modification. All lagging (hidden-state) dimensions are padded to a single global maximum rather than per-axis maxima, so that posterior vectors can be padded to the same length and contracted against any axis. This is necessary because marginalization involves such arbitrary contractions. 

Table~\ref{tab:variants} provides a summary of all possible approaches. Different variants arise depending on the merging strategy (axis-aligned or block-diagonal) and the memory optimization approach (BCOO sparse representation or tensor clustering).

\begin{table}[htbp]
\caption{Methods for optimization of message-passing algorithms.}
\label{tab:variants}
\centering
\begin{tabular}{|p{2.7cm}|p{1.8cm}|p{1.3cm}|p{1.1cm}|}
\hline
Method name & Merging\newline strategy\newline for A-arrays & Memory-saving technique & Log-likelihood deconstruction \\
\hline
Hybrid                  & Axis-aligned   & None       & $\checkmark$     \\
Sparse hybrid           & Axis-aligned   & BCOO       & $\checkmark$     \\
Clustered hybrid        & Axis-aligned   & Clustering & $\checkmark$     \\
\hline
Hybrid-block            & Block-diagonal & None       & $\checkmark$     \\
Sparse hybrid-block     & Block-diagonal & BCOO       & $\checkmark$     \\
Clustered hybrid-block  & Block-diagonal & Clustering & $\checkmark$     \\
\hline
End-to-end              & Axis-aligned   & None       & $\times$ \\
Sparse end-to-end       & Axis-aligned   & BCOO       & $\times$ \\
Clustered end-to-end    & Axis-aligned   & Clustering & $\times$ \\
\hline
\end{tabular}
\end{table}

We apply our optimization methodology to three message-passing algorithms: FPI, VMP, and MMP. It is worth noting that FPI computes only likelihood messages, whereas VMP and MMP require the calculation of both likelihood and transition messages. Nonetheless, since all three algorithms involve the likelihood messages, all approaches from Table~\ref{tab:variants} are applicable, and are empirically compared in section~\ref{sec:results}.

Among the nine approaches from Table~\ref{tab:variants}, selecting the best one for a given algorithm is not straightforward. Their relative performance depends on the POMDP specifications (see Section~\ref{sec:pomdp_configurations}), and benchmarking all nine variants before deployment is time-consuming, especially on embedded systems and when repeated across many POMDPs or hardware settings. To avoid this cost, we introduce an ML-based autotuner that takes a POMDP specification as an input and frames the selection of the best-performing method as a classification problem. Since FPI, VMP, and MMP correspond to different message-passing procedures, we instantiate this autotuning strategy separately for each algorithm family, rather than pooling them into a single classifier. In this work, we consider random forests (RF) and XGBoost as candidate classifiers. The details of the training data, fitting procedure, hyperparameters, and empirical results are presented in Section~\ref{sec:results}.


\section{Experimental Setup and Results}~\label{sec:results}

\vspace{-0.8cm}

\subsection{POMDP configurations}
\label{sec:pomdp_configurations}

To produce benchmark results that are broadly representative rather than tied to a small set of hand-crafted scenarios, we evaluate every implementation variant across the three message-passing algorithms (FPI, VMP, and MMP) on a large collection of randomly generated POMDP configurations. Random sampling exposes performance trends across heterogeneous tensor shapes and dependency structures that would be difficult to capture with only a handful of specific scenarios. A configuration is fully specified by the number of hidden state factors $F$, the number of observation modalities $M$, the per-factor and per-modality dimensionalities $\{n_1, n_2, ..., n_F\}$ and $\{m_1, m_2, ..., m_M\}$, and the A-dependency lists. The configuration generator works in three stages.

\subsubsection{Coarse dimensionality selection} The values of $F$ and $M$ are each drawn from the set $\{5, 10, 25, 125\}$, with the four levels acting as low, medium, high, and extreme numerical options. The upper limits on state and observation dimensionalities are drawn independently from $\{5, 10, 25\}$, with the lower limits fixed at $2$ (the lowest theoretically sensible value). We restrict our experiments to cases where $F \leq M$, as this corresponds to the most common setting in practice and ensures that each hidden state factor is associated with at least one observation modality. The choice of all numerical values was driven by our previous experience with real-world Bayesian applications and edge-cutting research in the literature \cite{saccade, automotive1, pezzato2025mobile}. In this manner, our generated POMDP configurations simulate realistic settings, while allowing for a high degree of variance that enriches statistical results.

\subsubsection{Fine-grained dimensionality sampling} Two complementary regimes for sampling the exact dimensions of each hidden state factor and observation modality are considered. A uniform regime samples each $n_j$ and $m_i$ independently and uniformly from their admissible ranges, i.e., starting at 2 and increasing to the corresponding upper bound, selected in the previous step. Conversely, a skewed regime samples roughly one-fifth of $n_j$ and $m_i$ variables from the top $10\%$ of their admissible ranges, and the remainder from the bottom $10\%$. This deliberately produces strongly imbalanced tensor shapes, serving as a further stress test of merged computations.

\subsubsection{A-dependency generation} Each hidden state factor is first guaranteed to appear in at least one A-dependency list, ensuring full coverage of the latent space with observation modalities. Afterward, the length $k_i$ of each list is sampled from $\{1, 2, ..., 10\}$, under an exponential prior that favors shorter lists. This follows a pattern frequently encountered in practice: a handful of observations can depend on a large number of hidden states (up to 10 in our case), but most observations usually depend on only a few states. After determining the lengths of A-dependency lists, the hidden-state factors that comprise each list are randomly drawn. Throughout this step, a negative correlation is enforced between each list length $k_i$ and the factor dimensions involved in it. Long dependency lists are biased toward low-dimensional factors, while short lists may involve any factor. This prevents a combinatorial explosion of A-array sizes that would be practically unfeasible to handle, while still aligning well with the structures commonly encountered in real-world Bayesian inference applications.

For each possible combination of high-level parameters determined in the first step, five independent POMDP configurations are sampled by repeatedly executing the second and third steps. This is done to account for the random noise in the generation procedure itself and to enhance the statistical significance of the main experimental results. The complete process yields $770$ POMDP specifications, which we utilize for experiments. Given a specification, the values in the A and B arrays are then randomly sampled from a uniform distribution and normalized to represent valid probability distributions. The full set of POMDP configurations with initialized A and B arrays is persisted to disk and reused across all subsequent experiments,
ensuring a fair comparison between various implementations of different algorithms.

\subsection{Benchmark results}

\begin{figure}[htbp]
\centering
\includegraphics[width=3.5in]{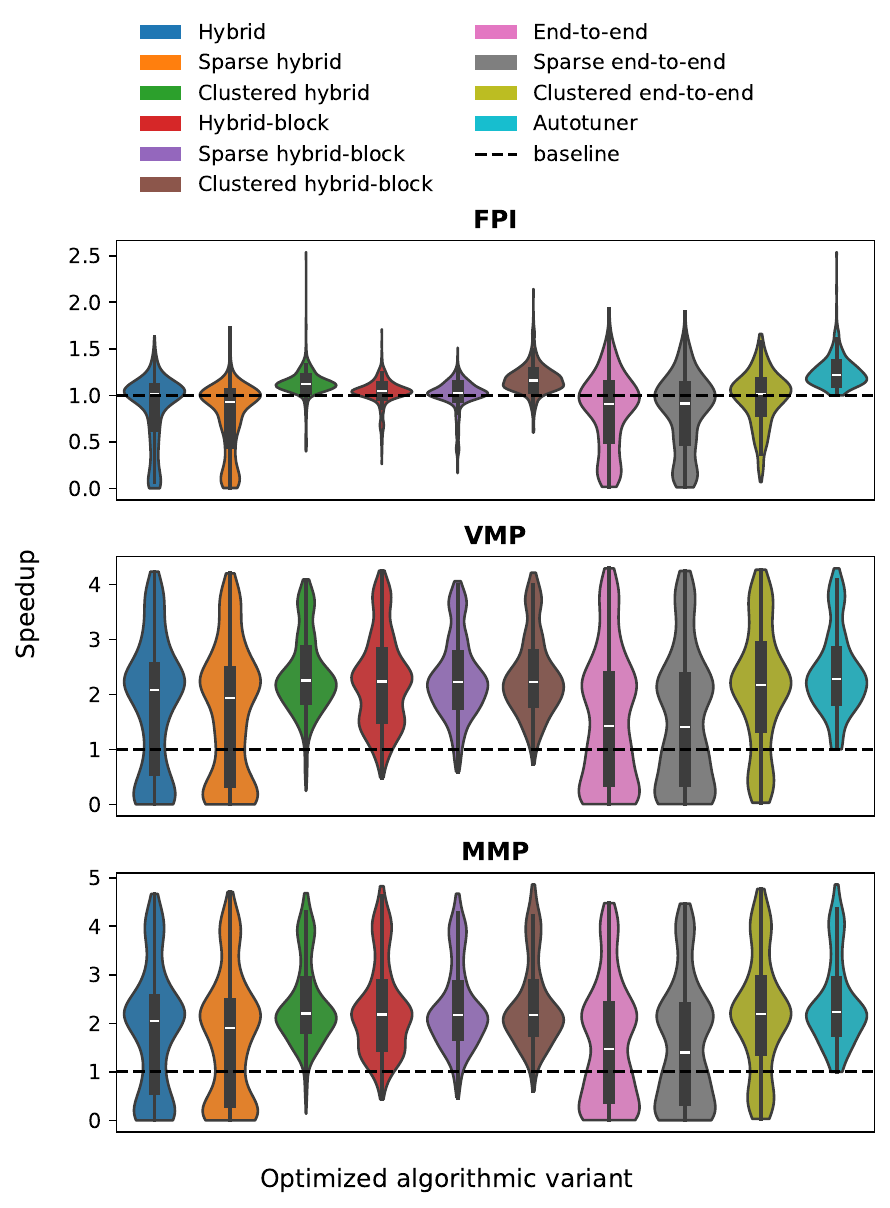}
\caption{Distributions of latency ratios (speedups) on an NV Jetson Orin AGX.} \vspace{-0.2cm}
\label{fig:latency_plots}
\end{figure}

Since our optimized variants of Bayesian inference algorithms are numerically equivalent to their original implementations, there is no need for accuracy-based evaluation. The experiments discussed in this section focus on measuring total inference latency, as it is the dominant deployment constraint on embedded platforms (e.g., in real-time applications, it determines whether a Bayesian algorithm has practical value). Each FPI, VMP, and MMP variant is benchmarked
on an NVIDIA Jetson Orin AGX and compared against the corresponding original implementation from
\texttt{pymdp}. This is repeated over the same persisted set of POMDP configurations described in Section~\ref{sec:pomdp_configurations}. We summarize the measurements using latency ratios: we divide the latencies of the original implementations (the baselines) by those of the optimized algorithm variants. Ratios above 1 represent speedups, and ratios below 1 denote slowdowns. Fig.~\ref{fig:latency_plots} shows a violin plot of these ratios over all POMDP configurations. Each violin corresponds to one optimized variant of FPI/VMP/MMP from Table~\ref{tab:variants}.

Some clear patterns emerge across the three algorithms. The speedups for FPI are the most modest, with most ratios falling below 1.5 and only a handful reaching 2-2.5. Some non-clustered hybrid and end-to-end implementations even incur slowdowns for a sizeable fraction of POMDP configurations. This is largely expected, as FPI relies solely on likelihood messages evaluated at a single time step, offering relatively little regular structure for our merged contractions to exploit. Nonetheless, the clustered hybrid method, as well as the hybrid-block variants, yield consistent speedups.

In contrast to the FPI algorithm, VMP and MMP benefit much more from our methodology. The best-performing custom variants achieve average speedups of around 2.5, peaking at 4.3 for VMP and 4.9 for MMP. For both algorithms, such good results are achieved for two main reasons. First, optimized implementations accelerate the computation of both transition and likelihood messages, yielding better latency savings than FPI. Second, they operate over a temporal window, which our methodology parallelizes effectively. The resulting increase in regular, batchable work maps precisely and efficiently onto the GPU. It is worth emphasizing that MMP, arguably the most biologically plausible and most widely used of the three algorithms in practice, exhibits the largest peak gains. Conversely, the weaker FPI results are of
limited practical concern given that FPI is by far the cheapest of the three
algorithms in absolute terms.

Comparing the two core families of optimized implementations, we observe that the end-to-end methods generally perform worse than the hybrid variants, but yield slightly higher upper tails, i.e., larger speedups. However, this also comes at the cost of much heavier lower tails. Uniform padding to a global maximum dimensionality introduces additional redundant work that often offsets the benefit of avoiding intermediate tensor deconstruction. Overall, the average speedups of the end-to-end methods are consistently lower than those of the hybrid families.

Next, we contrast the two merging strategies (Section~\ref{sec:optimizing_contractions}) between the hybrid families of methods. A consistent robustness advantage of the block-diagonal merging is evident. The hybrid-block methods achieve speedups across the vast majority of POMDP configurations, with only a negligible fraction experiencing slowdowns. While the axis-aligned hybrid variants achieve competitive average speedups, their full distributions exhibit pronounced lower tails in non-clustered settings, indicating frequent slowdowns on unfavorable input shapes. These performance differences among the merging strategies can be attributed to the block-diagonal layout's core feature. It reduces the volume of padded zeros for the heterogeneous, high-rank A-arrays, yielding higher performance across the space of possible POMDP configurations.

The two memory-saving mechanisms behave very differently with respect to latency. Introducing BCOO sparse representations does not, on its own, translate into noticeably faster execution. The violin plots show that the sparse methods achieve performance largely equivalent to that of their non-sparse counterparts. This can be explained by JAX's current, very limited, and experimental support for sparse arrays. Nonetheless, BCOO sparse representations help keep the parameter count of merged contractions down, reducing it by 14x on average. Once highly optimized sparse operations are available, it might be worthwhile to revisit this aspect of our work and seek BCOO-related speedups. Conversely, the clustering approach not only reduces the parameter count for padded arrays by approximately 9x, but also accelerates inference. The clustered variants attain tighter distributions of latency ratios, thinning out the lower tails remarkably well, and thus leading to the highest average speedups for every algorithm. Clustering minimizes wasted time on padded zero entries while enhancing the hardware benefits of compact, merged operations.




In the rightmost distribution in each panel of Fig.~\ref{fig:latency_plots}, we can observe the autotuner results. The autotuner selects the algorithmic variant expected to be the fastest for each POMDP configuration. Because the optimal variant is generally configuration-dependent, no single fixed method dominates everywhere, and the autotuner consequently tracks the upper envelope of all variants. Its latency ratio distribution is shifted upward, and it almost never drops below the baseline. This confirms the practical value of configuration-aware selection, enabling a deployed Bayesian algorithm to achieve near-best-case acceleration without the cost of benchmarking every variant at deployment time. Taken together, these results show that our methodology delivers consistent speedups across distinct message-passing algorithms and that a lightweight autotuner suffices to automatically realize the benefit.

\input{autotuner_table}
In more detail, to develop an autotuner for each algorithm (FPI/VMP/MMP), we use the latency measurements collected across the $770$ benchmarked POMDP configurations. We adopt a cross-validation scheme in which, for each fold, $80\%$ of the configurations are used for training, and the remaining $20\%$ are held out for testing. The training split is further split internally for model selection, which is performed via a randomized search over $60$ hyperparameter configurations.
The most relevant searched hyperparameters include the number of estimators and maximum tree depth for both RF and XGBoost, the learning rate and $L_1$/$L_2$ regularization terms for XGBoost, and the minimum number of samples required for node splitting and leaf creation for RF. A summary of all the considered hyperparameters is shown in Table~\ref{tab:hyperparams}. We evaluate the quality of the autotuner using a \textit{regret} metric, defined as the mean latency degradation incurred when selecting the configuration with the highest classifier logit relative to the true best-performing configuration identified by the oracle benchmark. The resulting regret values vary across inference algorithms and classifier families. For FPI, XGBoost, and RF obtain regrets of $3.15\%$ and $3.6\%$, respectively. For VMP, the corresponding regrets are $1.64\%$ and $4.78\%$, while for MMP they are $1.58\%$ and $1.94\%$. XGBoost is consistently better across all three algorithm families, and is therefore the classifier used for the autotuner results reported above.

\begin{figure}[htbp]
\vspace{-0.1cm}
\centering
\includegraphics[width=3.5in]{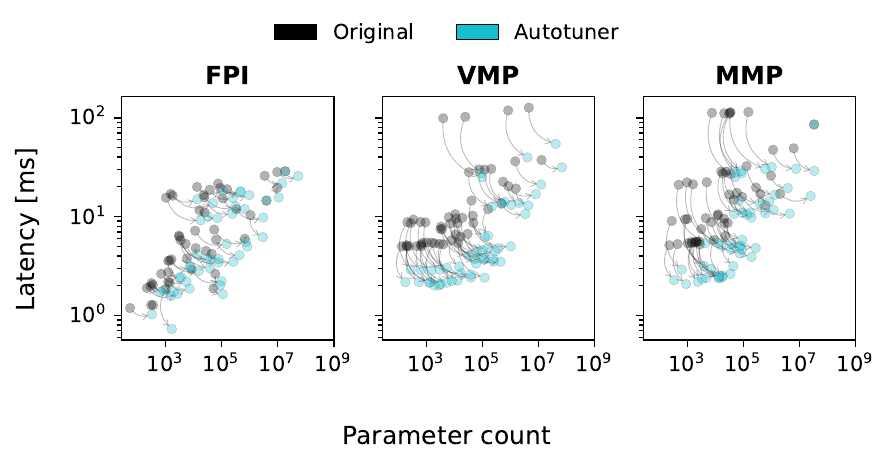}\vspace{-0.3cm}
\caption{A comparison between parameter counts and latencies of original algorithms and the best custom variants. A pair of points connected by an arrow represents a single POMDP configuration.} \vspace{-0.3cm}
\label{fig:memory_plots}
\end{figure}

Finally, we outline the trade-off between parameter counts and latencies in the proposed methodology. We opt not to show physical memory measurements since they can be inconsistent and unreliable, due to being dependent on versions of software libraries and the specifics of a given hardware platform. Instead, the parameter count represents a universal, analytically computable metric, that is in direct correlation to actual memory consumption. Fig.~\ref{fig:memory_plots} shows that reductions in latency may sometimes lead to slight increases in parameter counts. However, since such changes generally translate to memory differences of less than 1GB, we argue that this is a completely acceptable trade-off in GPU-oriented edge applications, where several GB of VRAM are usually available. To avoid graph overcrowding, we compare only the best custom methods (as predicted by the autotuner) with the baselines across a uniform sample of 50 POMDP configurations.


\section{Conclusion and Future Work}

In this work, we introduced a hardware-oriented methodology for accelerating discrete Bayesian inference on embedded GPUs. By identifying tensor contractions as the computational primitive shared across a wide range of variational message-passing algorithms, we were able to formulate optimizations at the level of these primitives, yielding generalized inference speed-ups across algorithms. The proposed tensor merging strategies, together with the memory-saving techniques, replace irregular loops of small contractions with compact operations that map efficiently onto GPU hardware. Instantiating this methodology in optimized FPI, VMP, and MMP variants and benchmarking them on an NVIDIA Jetson Orin AGX across $770$ POMDP configurations, we demonstrated speedups of up to 5x without any loss of numerical accuracy. We further showed that a lightweight, ML-based autotuner can recover most of the available acceleration automatically, while avoiding the cost of exhaustive per-deployment benchmarking.

Several directions remain open for future work. The current methodology supports arbitrary A-dependencies but is restricted to trivial B-dependencies. Extending the merging strategies to arbitrary transition structures would broaden their applicability to a richer set of models. Our use of sparse BCOO arrays is presently constrained by the experimental state of sparse operations in JAX, and is likely to yield additional speedups as these backends mature. Finally, our evaluation is confined to a single embedded platform due to the high time-consuming nature of the experiments. Benchmarking across a more diverse set of hardware targets would further strengthen the practical case for the proposed optimizations.

\section*{Acknowledgments}


The authors used Anthropic's Claude (Opus 4.8) to assist with the organization and structuring of the manuscript. However, there are no purely generated parts, as all text was carefully edited, reviewed, and verified by the authors, who take full responsibility for the content.

{
\appendix[]
\label{app:implementation}

This appendix provides the concrete construction details for the methodology outlined in Section~\ref{sec:methodology}. We specify how the merged arrays are built for the axis-aligned and block-diagonal strategies, how tensor contractions can be performed in merged form, and how the methodology extends to temporal sequences, as required by the VMP and MMP algorithms.

\subsection{Constructing merged tensor contractions}

We use the log-likelihood computation introduced in Section~\ref{sec:optimizing_contractions} as a running illustrative example, since it instantiates both merging strategies in their most general form.

In the axis-aligned padding and merging approach, the tensor $\mathbf{\bar{A}}$ is constructed as follows. Its rank (i.e., the number of axes) is equal to the largest rank among the individual A-arrays: $rank(\mathbf{\bar{A}}) = \underset{1 \leq i \leq M}{\max} (rank(A_i))$. This rank comprises one batch axis, one observation axis, and $k$ hidden-state axes, also referred to as lagging dimensions in the literature. The number $k$ is equal to the length of the longest A-dependency list in a given setting. The batch axis of $\mathbf{\bar{A}}$ needs to have a length of $M \cdot b$, to account for all $M$ original arrays. The length of the observation axis is chosen as the maximum over the numbers of levels of all observation modalities, i.e., $\underset{1 \leq i \leq M}{\max} m_i$. The remaining axes, representing the hidden state factors, are treated similarly. Overall, the shape of $\mathbf{\bar{A}}$ becomes $Mb \times \underset{1 \leq i \leq M}{\max} m_i \times \underset{1 \leq i \leq M}{\max} n_{i_1} \times ... \times \underset{1 \leq i \leq M}{\max} n_{i_k}$. The array $\mathbf{\bar{A}}$ is first initialized with zero values, and afterwards we iterate over original arrays $A_i$ and incorporate their values: $\mathbf{\bar{A}}[b(i - 1):bi, 0:m_i, 0:n_{i_1}, ..., 0:n_{i_{k_i}}] = A_i$. The result is therefore a zero-padded concatenation of $\{A_1, A_2, ..., A_M\}$ along the batch axis, as depicted in Fig.~\ref{fig:padding_and_merging}. The observation vectors $o^i$ are merged in the same manner, producing a zero-padded matrix $\mathbf{\bar{o}}$ of shape $Mb \times \underset{1 \leq i \leq M}{\max} m_i$. To match the rank of $\mathbf{\bar{A}}$, this matrix is broadcast to shape $Mb \times \underset{1 \leq i \leq M}{\max} m_i \times \underbrace{1 \times 1 \times ... \times 1}_{k}$. The log-likelihoods in merged form are then computed by element-wise multiplication of $\log(\mathbf{\bar{A}})$ and the broadcast $\mathbf{\bar{o}}$, followed by summing out the observation axis (Fig.~\ref{fig:merged_computation}). The merged output can be deconstructed into a list of per-modality outputs (Fig.~\ref{fig:deconstruction}) when required by downstream computations, e.g., in the hybrid implementations of FPI, VMP, and MMP, as noted in Section~\ref{sec:variants}.

When using axis-aligned merging in the end-to-end algorithmic variants, a slight modification is introduced. Each lagging dimension of $\mathbf{\bar{A}}$ gets padded to a global maximum $n_{max} = \underset{1 \leq j \leq F}{\max} n_j$, rather than keeping track of per-axis maximum lengths. This yields a merged array of shape $Mb \times \underset{1 \leq i \leq M}{\max} m_i \times \underbrace{n_{max} \times n_{max} \times ... \times n_{max}}_{k}$. After padding the posterior approximation vectors to $n_{max} = \underset{1 \leq j \leq F}{\max} n_j$ as well, arbitrary contractions between them and $\mathbf{\bar{A}}$ become possible. This is crucial for the marginalization step when computing likelihood messages, and it enables end-to-end implementations of FPI, VMP, and MMP to use padded and merged tensors throughout the process.

For block-diagonal merging, we first reshape each A-array. For any index $i$, the lagging dimensions of $A_i$ are flattened into a single dimension. In the unbatched case, this turns an array of shape $m_i \times n_{i_1} \times n_{i_2} \times ... \times n_{i_{k_i}}$ into a $p_i \times m_i$ matrix, where $p_i$ represents the flattened product of lagging dimension lengths: $p_i = n_{i_1} \cdot n_{i_2} \cdot ... \cdot n_{i_{k_i}}$. The reshaped A-arrays are then arranged into one large block-diagonal matrix $\mathbf{A_{block}}$, of shape $\left( \underset{1 \leq i \leq M}{\sum} p_i \right) \times \left( \underset{1 \leq i \leq M}{\sum} m_i \right)$. The observation vectors $o^i$ are concatenated into one long vector $\mathbf{o_{concat}}$ with $\underset{1 \leq i \leq M}{\sum} m_i$ entries. Afterward, the contraction operation reduces to a standard matrix-vector multiplication $\log(\mathbf{A_{block}}) \cdot \mathbf{o_{concat}}$, as illustrated in Fig.~\ref{fig:block_computation}. The resulting vector can be decomposed into a list of arrays that would be produced by individual contractions. This involves splitting it into parts of $p_1, p_2,...,p_M$ elements respectively, and reshaping the $i$-th part into an $n_{i_1} \times n_{i_2} \times ... \times n_{i_{k_i}}$ array. When computations need to be performed in batched form (i.e., with A-arrays of original shapes $b \times m_i \times n_{i_1} \times ... \times n_{i_{k_i}}$ and observations represented as $b \times m_i$ matrices), the merged operation $\mathbf{A_{block}} \cdot \mathbf{o_{concat}}$ is implemented either via a batched \texttt{matmul}, a more general \texttt{einsum}, or by simply applying a \texttt{vmap} over the batch axis.

\subsection{Handling temporal sequences}

So far, we have provided mathematical examples without a time dimension for simplicity. In the case of VMP and MMP algorithms, observations and hidden states are considered for multiple consecutive time steps. Our methodology handles such cases by applying a \texttt{vmap} over the time axis, and keeping the rest of the logic as-is. This is a simple yet effective choice that avoids additional mathematical and technical complications and is well aligned with our hardware-aware optimization objectives. By \texttt{vmapping} over a time axis of length $h+1$ (spanning the inference horizon $h$ and the current time step), we essentially evenly distribute merged tensor contractions into $h+1$ parallel threads, as long as the given hardware constraints allow for that.

}


\bibliographystyle{IEEEtran}
\bibliography{references}

\vfill

\end{document}

%% file: autotuner_table.tex
\begin{table}[ht]
\centering
\caption{Hyperparameter search space for the RF and XGBoost models.}
\label{tab:hyperparams}
\resizebox{\columnwidth}{!}{%
{
\begin{tabular}{lll}
\hline
\textbf{Model} & \textbf{Hyperparameter} & \textbf{Values considered} \\
\hline
Random Forest & \texttt{n\_estimators} & 300, 500, 700, 900, 1200 \\

Random Forest & \texttt{max\_depth} & 6, 8, 10, 12, 16, 20, None \\

Random Forest & \texttt{min\_samples\_split} & 2, 4, 6, 8, 10, 14, 18 \\

Random Forest & \texttt{min\_samples\_leaf} & 1, 2, 3, 4, 5, 7, 10 \\

Random Forest & \texttt{max\_features} & sqrt, log2, 0.3, 0.5 \\

Random Forest & \texttt{max\_samples} & None, 0.7, 0.8, 0.9 \\

Random Forest & \texttt{min\_impurity\_decrease} & 0.0, $10^{-4}$, $5 \times 10^{-4}$, $10^{-3}$ \\

Random Forest & \texttt{class\_weight} & None, balanced \\
\hline
XGBoost & \texttt{n\_estimators} & 100, 300, 500, 700, 1000, 1500 \\

XGBoost & \texttt{max\_depth} & 3, 4, 5, 6, 7, 8, 10 \\

XGBoost & \texttt{learning\_rate} & 0.01, 0.03, 0.05, 0.1, 0.2 \\

XGBoost & \texttt{subsample} & 0.6, 0.7, 0.8, 0.9, 1.0 \\

XGBoost & \texttt{colsample\_bytree} & 0.6, 0.7, 0.8, 0.9, 1.0 \\

XGBoost & \texttt{min\_child\_weight} & 1, 3, 5, 7, 10 \\

XGBoost & \texttt{gamma} & 0, 0.1, 0.2, 0.3, 0.5 \\

XGBoost & \texttt{reg\_alpha} & 0, 0.01, 0.1, 1 \\

XGBoost & \texttt{reg\_lambda} & 1, 2, 5, 10 \\
\hline
\end{tabular}
}
}
\end{table}